\documentclass[fleqn,10pt]{wlscirep}
\usepackage[utf8]{inputenc}
\usepackage[T1]{fontenc}
\usepackage{amsmath}
\usepackage{amssymb}
\usepackage[table]{xcolor}
\usepackage{tabularx}
\usepackage{array}
\usepackage{placeins}

\title{CoMNet: A MedNeXt-CorrDiff Framework for Multi-Site Brain Tumor Segmentation}

\author[1]{Michael L. Evans}
\author[1]{MD Fayaz Bin Hossen}
\author[1]{MD Shibly Sadique}
\author[1]{Walia Farzana}
\author[1,*]{Khan M. Iftekharuddin}

\affil[1]{Department of Electrical and Computer Engineering, Old Dominion University, Norfolk, VA, USA}

\affil[*]{Corresponding author: kiftekha@odu.edu}

\keywords{glioma segmentation, UTSW-Glioma, magnetic resonance imaging, MedNeXt, CorrDiff, model ensembling, medical image analysis}

\begin{abstract}
Accurate brain tumor segmentation from multiparametric magnetic resonance imaging (MRI) is critical for treatment planning, response assessment, and neuro-oncology research. However, automated segmentation remains a difficult task in computer vision because of variation in tumor appearance and MRI protocols across patient scans. Moreover, clinically important regions such as enhancing tumor and tumor core are often small relative to the full brain volume, further increasing the difficulty of achieving high voxel-level precision. These challenges are amplified in multi-site datasets, where differences in scanner hardware and acquisition parameters can introduce non-biological variation. To address this, networks must learn tumor-specific features while remaining robust to site-dependent noise. In this paper, we show that an ensemble of multi-fold predictions from a modern 3D convolutional segmentation network with corrective diffusion (CorrDiff) post-processing improves brain tumor segmentation across datasets. We propose CoMNet, an ensembled MedNeXt-CorrDiff framework for accurate multi-site brain tumor segmentation. In this framework, we use MedNeXt as the primary segmentation model for feature learning, while a corrective diffusion block learns to refine the residual errors in the individual prediction maps before probabilistic thresholding. This process reduces the variance across fold predictions by correcting fold-specific residual errors and aggregating them into a consensus mask that is less sensitive to site-dependent imaging variability. Our proposed framework achieved the highest Dice score compared to two baseline models on the UTSW-Glioma and BraTS-SSA datasets. Experimental results support the use of corrective diffusion and fold-level probability ensembling as meaningful additions to existing state-of-the-art models for accurate glioma segmentation on multi-site datasets.
\end{abstract}
\begin{document}

\flushbottom
\maketitle
\thispagestyle{empty}

\section{Introduction}

Gliomas are among the most clinically significant brain tumors, and their management depends heavily on the accurate interpretation of magnetic resonance imaging (MRI). Manual labeling of the tumor regions is time consuming, has high variability across experts, and is difficult to scale across larger studies. Accurate segmentation is critical for effective tumor resection, and plays a significant role in the accuracy of diagnosis and treatment planning. Automated brain tumor segmentation therefore remains an important problem in medical image analysis, particularly for quantitative assessment of the enhancing tumor (ET), tumor core (TC), and whole tumor (WT) regions. In recent years, considerable progress has been made in image processing for clinical tasks, led by advances in representation learning and algorithm design \cite{suganyadevi2022review, chan2020deep, liu2023deep}. However, maintaining accurate and reliable segmentation across heterogeneous multi-site MRI datasets remains difficult, motivating continued development of robust automated segmentation frameworks \cite{pemberton2023multi, bento2022deep, jiang2024enhancing}.

Beyond random and systematic segmentation errors, multi-site MRI data introduce an additional challenge related to model generalizability. Images acquired across different institutions often vary in scanner hardware, acquisition parameters, spatial resolution, intensity distribution, and patient demographics \cite{Adewole2025}. As a result, a segmentation model may learn features that are partly dependent on site-specific imaging characteristics rather than tumor-specific structure, leading to reduced performance when applied to data from new centers or cohorts. This problem is especially relevant for glioma segmentation, where the enhancing tumor, tumor core, and whole tumor regions differ in volume, morphology, and contrast appearance. Therefore, developing automated segmentation frameworks that are robust to site-dependent variation and capable of maintaining reliable performance across heterogeneous MRI datasets remains an important goal in medical image analysis. 

To improve robustness to variability, glioma segmentation methods commonly rely on multiparametric MRI, where complementary imaging sequences provide distinct but related information about tumor structure and surrounding tissue \cite{de2024modality}. Pre-contrast T1-weighted imaging (T1n) provides anatomical structure, post-contrast T1-weighted imaging (T1c) highlights regions of blood-brain barrier disruption and enhancement, T2-weighted imaging (T2w) captures fluid-sensitive abnormality, and T2-FLAIR suppresses cerebrospinal fluid to improve visualization of edema. In this study, we use these four modalities to predict standardized BraTS-style tumor regions, allowing the model to distinguish clinically relevant subregions.

Although Transformer-based models have gained increasing attention, convolutional encoder-decoder networks remain widely used backbones for data-limited medical imaging tasks because their spatial inductive bias is well suited for dense prediction. ConvNeXt helped bridge these two design directions by showing that a pure convolutional network could incorporate Transformer-inspired principles while retaining the efficiency and locality of convolutional operations \cite{Liu2022}. These changes focused on large-kernel depthwise convolutions, inverted bottlenecks, residual connections, and improved normalization. MedNeXt adapts this design philosophy to 3D medical image segmentation by integrating ConvNeXt-style blocks into a volumetric encoder-decoder architecture and scaling the model across depth, width, and kernel size \cite{Roy2023,Roy2025MedNeXtV2}. This makes MedNeXt an effective backbone for glioma segmentation, where the model must capture local boundary detail, multi-modal MRI contrast, and broader volumetric context. Deep learning approaches to automatic brain tumor segmentation often enhance the CNN backbone with post-processing to refine the prediction errors introduced by the segmentation model \cite{Li2024}. Recently, the corrective diffusion process has become a popular method in image segmentation, wherein noise is gradually added to an image, and the network is trained to reverse this process. In the context of segmentation, corrective diffusion can learn to correct the systematic errors between the predicted mask and the manual segmentation. Since the unique error map is dependent on the network making the prediction, we combine this process with post hoc ensembling, where multiple models are combined and averaged to improve the initial prediction mask. 

Heterogeneity in multi-site MRI, caused by differences in scanner field strength, acquisition protocols, reconstruction parameters, spatial resolution, and intensity distributions, presents a substantial challenge for model training and deployment \cite{bento2022deep, guan2021domain, kushol2023dsmri}. These site-dependent differences can introduce non-biological variation into the image appearance, making it difficult for segmentation models to separate tumor-specific features from scanner- or protocol-specific patterns \cite{bento2022deep, liu2024learning}. Traditional harmonization approaches attempt to reduce this variability before model training by standardizing image intensities, normalizing acquisition-related effects, or mapping images into a shared feature or appearance space \cite{abbasi2024deep, liu2024learning}. More recent learning-based and diffusion-based harmonization methods extend this idea by modifying image appearance across domains to reduce scanner- or site-specific differences prior to downstream segmentation \cite{durrer2023diffusion, wu2025unpaired, lan2025diffusion}. Although these approaches can reduce imaging variability, they also introduce additional preprocessing complexity and may alter intensity patterns that are relevant for tumor delineation. This is particularly important in glioma segmentation, where subtle contrast differences help distinguish enhancing tumor, necrotic core, edema, and surrounding tissue.

These limitations motivate an alternative strategy that improves robustness at the prediction level rather than modifying the input MRI scans directly. In this work, corrective diffusion is used to refine model-specific segmentation errors, while fold-level ensembling reduces prediction variance across independently trained models. Together, these components provide a harmonization-free refinement strategy for improving segmentation consistency across heterogeneous MRI data. In this work, we propose CoMNet, a MedNeXt-CorrDiff framework for multi-site glioma segmentation on the UTSW-Glioma and Brain Tumor Segmentation in Sub-Saharan Africa (BraTS-SSA) datasets. The framework uses MedNeXt as the primary segmentation model, CorrDiff for corrective diffusion post-processing, and fold-level ensembling to combine the predictions of independently trained cross-validation models. We compare CoMNet against SegResNet and standalone MedNeXt to show that our proposed refinement and ensembling strategy improves segmentation performance beyond two strong baselines. The main contributions of this work are as follows:

\begin{itemize}
\item We evaluate automated glioma segmentation on the UTSW-Glioma and BraTS-SSA datasets using four MRI modalities and BraTS-style enhancing tumor, tumor core, and whole tumor labels.
\item We integrate CorrDiff as a corrective diffusion post-processing component that refines residual segmentation errors from MedNeXt probability maps.
\item We apply fold-level ensembling and probabilistic thresholding so that predictions from independently trained cross-validation models contribute to the final segmentation mask.
\item We show that CoMNet improves held-out test set Dice score compared with SegResNet and standalone MedNeXt across multiple glioma MRI datasets.
\end{itemize}

\section{Related works}

Most modern medical image segmentation pipelines build on the encoder-decoder structure first introduced by U-Net, where a contracting path learns semantic context and an expanding path restores spatial resolution through skip connections \cite{Ronneberger2015}. This design is especially well suited for dense prediction because it combines high-level semantic features with lower-level spatial detail. For volumetric medical imaging, 3D U-Net and V-Net extended this idea by processing full 3D image volumes rather than treating each 2D slice independently \cite{Cicek2016, Milletari2016}. These volumetric extensions benefit MRI-based tumor segmentation because glioma appearance is inherently three-dimensional, and the anatomical relationships across adjacent slices provide useful context for defining tumor boundaries.

\begin{figure}[!t]
\centering
\includegraphics[width=\columnwidth]{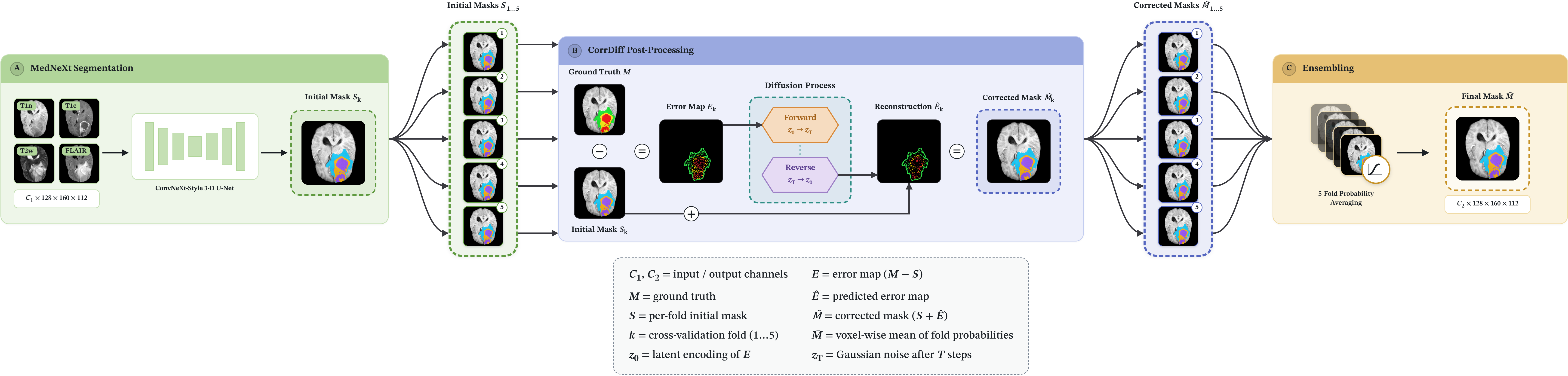}
\caption{Overview of the proposed CoMNet pipeline. Four MRI modalities are preprocessed and passed through the MedNeXt segmentation backbone. CorrDiff is then used as a corrective diffusion module for residual error refinement. Predictions from the unique fold-specific models are ensembled before final evaluation using BraTS-style tumor subregions: enhancing tumor (ET), tumor core (TC), and whole tumor (WT).}
\label{fig:pipeline}
\end{figure}

Building on this encoder-decoder foundation, residual convolutional networks have become strong baselines for brain tumor segmentation and related volumetric segmentation tasks. SegResNet follows the same general multi-resolution structure but incorporates residual blocks to improve gradient flow and feature reuse in deeper networks \cite{Myronenko2018}. This is inherently useful for glioma segmentation to allow the model to preserve low-level boundary information while also learning higher-level tumor context across the volume. In this work, we include SegResNet as an established 3D residual segmentation baseline for comparison with CoMNet.

The continued success of convolutional encoder-decoder models was further demonstrated by nnU-Net, which later showed that a carefully configured U-Net can remain highly competitive across a wide range of medical segmentation tasks when preprocessing, patch size, network configuration, and training settings are adapted to the dataset \cite{Isensee2021}. At the same time, Transformer and hybrid architectures have been increasingly applied to medical image segmentation because attention mechanisms can model long-range dependencies \cite{Chen2021,Hatamizadeh2021,Wang2021}. However, fully volumetric Transformer-style models can require substantial GPU memory and often benefit from larger annotated datasets, which are less common in medical imaging than in natural image domains. These constraints motivate continued interest in modernized convolutional architectures that retain the efficiency and spatial inductive bias of CNNs while improving their ability to capture broader context.

ConvNeXt was motivated by the idea that the apparent advantage of Transformer-based vision models was partly tied to architectural and training choices rather than the attention mechanism itself. By progressively modernizing a ResNet-style architecture, ConvNeXt demonstrated that pure convolutional networks can benefit from large-kernel depthwise convolutions, inverted bottlenecks, fewer activation and normalization layers, and Transformer-inspired scaling while preserving the efficiency and locality of convolutional operations \cite{Liu2022}. This design direction is particularly relevant for medical imaging, where annotated datasets are often limited and fully volumetric models must balance representational capacity with computational feasibility.

MedNeXt adapts this ConvNeXt design philosophy to 3D medical image segmentation. Rather than using ConvNeXt as a classification backbone, MedNeXt places 3D ConvNeXt-style blocks within an encoder-decoder segmentation architecture \cite{Roy2023}. This preserves the main strengths of U-Net-like models, including multi-resolution feature learning and skip connections, while replacing conventional convolutional blocks with modernized ConvNeXt blocks. MedNeXt also introduces residual upsampling and downsampling blocks to preserve semantic information as the feature maps change spatial resolution. These properties make MedNeXt well suited for glioma segmentation, where the network must capture small enhancing regions, tumor core boundaries, multi-modal MRI contrast, and broader whole-tumor context within the same volumetric prediction.

Recent extensions to ConvNeXt have further improved convolutional representation learning. ConvNeXt V2 introduced Global Response Normalization (GRN) to improve feature competition across channels and reduce redundant feature responses \cite{Woo2023}. MedNeXt-v2 extends this idea to 3D medical image segmentation by incorporating 3D GRN and scaling the architecture for volumetric representation learning \cite{Roy2025MedNeXtV2}. In this paper, we refer to the updated GRN-based version of MedNeXt-v2 simply as MedNeXt. Our motivation for using MedNeXt is that it brings ConvNeXt-style large-context feature learning into a practical 3D encoder-decoder framework for medical image segmentation.

Alongside advances in segmentation backbones, diffusion models have become increasingly popular in medical imaging. Diffusion models learn to reverse a gradual noising process and have been applied to image generation, reconstruction, and segmentation \cite{Ho2020, Kazerouni2023}. In segmentation, many diffusion-based methods generate a segmentation mask directly from the input image, which can be computationally expensive and may replace an otherwise strong discriminative segmentation model \cite{Wolleb2022, Wu2024}. CorrDiff takes a different approach by using corrective diffusion as a post-processing step rather than as the primary segmentation model \cite{Li2024}. Instead of predicting the full segmentation mask from scratch, CorrDiff learns to correct systematic residual errors between an initial model prediction and the corresponding manual annotation. The original CorrDiff framework used a U-Net segmentation backbone; in this work, we replace that backbone with MedNeXt to produce stronger initial probability maps. CorrDiff is then applied to refine the residual errors in the MedNeXt predictions, allowing the final framework to combine a modern 3D convolutional backbone with diffusion-based segmentation correction.

\section{Methodology}

In this paper, the CoMNet framework uses three steps for 3D brain tumor segmentation, shown in Figure \ref{fig:pipeline}. Each subject is loaded with four MRI modalities and a segmentation mask. The volumes are standardized through orientation correction, nonzero-voxel normalization, and fixed-size cropping or padding. MedNeXt is trained as the primary segmentation backbone, and CorrDiff then refines the initial prediction mask by modeling the residual error. Finally, fold-specific predictions are ensembled at the probability-map level and thresholded to produce the final ET, TC, and WT masks.

\subsection{Datasets and preprocessing}

We evaluate our model on brain tumor datasets from the Brain Tumor Segmentation (BraTS) Challenge \cite{menze2014multimodal} and the University of Texas Southwestern (UTSW) Medical Center \cite{Reddy2024,Reddy2026}. The UTSW-Glioma dataset is a curated glioma multi-MRI dataset that is publicly available through The Cancer Imaging Archive. The dataset contains 625 patients treated at UTSW between 2006 and 2023. Each patient record includes pre-contrast T1-weighted, post-contrast T1-weighted, T2-weighted, and T2-FLAIR images. This dataset also includes demographic, histopathologic, and molecular marker information, including IDH mutation status, 1p/19q codeletion status, MGMT promoter methylation status, tumor type, and tumor grade. These molecular and clinical variables are not used as model inputs in this study, because our goal is segmentation from MRI. However, they may make the dataset valuable for future multimodal analysis of segmentation performance with clinical data.

This study also utilizes the BraTS-SSA dataset \cite{Adewole2025}, which comprises of 60 preoperative glioma patients from Sub-Saharan Africa. Each case includes four MRI sequences: T1-weighted, T1-Contract Enhanced, T2-weighted, and T2-Fluid Attenuated Inversion Recovery. This dataset represents a patient population and imaging setting that are underrepresented in many brain tumor segmentation benchmarks. Compared with larger multi-institutional datasets, BraTS-SSA introduces additional challenges related to limited sample size, heterogeneous clinical acquisition protocols, and high variation in image quality across scans. These challenges are especially relevant for automated segmentation because differences in contrast, resolution, tumor presentation, and scanner protocols can affect the visibility of tumor boundaries and subregions.

Each subject is required to include the four MRI modalities and a corresponding segmentation mask to serve as the ground truth. The segmentation labels are converted into three binary tumor-region masks: ET corresponds to enhancing tumor, TC includes the enhancing and core tumor components, and WT includes the full visible tumor, including edema. This conversion gives the model three output channels and keeps the evaluation focused on clinically meaningful tumor regions. Before training, each subject directory is checked and validated for all required images and the corresponding segmentation mask. Subjects with missing modalities, missing masks, unreadable files, or incompatible image geometry are excluded from training. All volumes are converted to a common canonical orientation. Each modality is then normalized using z-score normalization over nonzero voxels:
\begin{equation}
\hat{x}_{i} = \frac{x_i - \mu_{\Omega}}{\sigma_{\Omega} + \epsilon},
\end{equation}
where $x_i$ is the intensity at voxel $i$, $\Omega$ is the set of nonzero voxels for that modality, $\mu_{\Omega}$ and $\sigma_{\Omega}$ are the mean and standard deviation over nonzero voxels, and $\epsilon$ is a small constant for numerical stability. Normalizing only nonzero voxels prevents the background from corrupting the estimated MRI intensity data distribution. After normalization, volumes are cropped to a fixed training patch size of $128 \times 160 \times 112$. This patch size preserves sufficient anatomical context while allowing 3D training on our available GPU memory. During training, we use lightweight spatial and intensity augmentation, limited to random flips, random affine transformations, random gamma adjustment, random noise, and random blur. Augmentation is applied only during training, validation inference uses deterministic preprocessing.

\subsection{CoMNet framework}

CoMNet is designed as a multi-component segmentation pipeline rather than an architectural modification to existing state-of-the-art models. The proposed framework consists of three core components: a MedNeXt 3D segmentation backbone, a CorrDiff corrective diffusion module, and a fold-level probability ensemble. For each cross-validation fold, MedNeXt first generates volumetric probability maps that provide the initial segmentation mask. CorrDiff then refines these probability maps by learning to correct residual segmentation errors between the prediction and the corresponding manual annotation. Finally, the corrected predictions from independently trained fold models are combined through probability ensembling and thresholded to produce a more stable final segmentation mask.

\begin{table*}[!t]
\centering
\begingroup
\renewcommand{\arraystretch}{1.25}
\setlength{\tabcolsep}{8pt}
\renewcommand{\tabularxcolumn}[1]{m{#1}}
\begin{tabularx}{\textwidth}{|>{\raggedright\arraybackslash}X|
                                >{\raggedright\arraybackslash}X|}
\hline
\rowcolor{gray!15}
\textbf{Setting} & \textbf{Value} \\
\hline
Segmentation backbone & 3D MedNeXt-B with GRN \\
\hline
MedNeXt loss & DSC / DSC++ with binary cross-entropy \\
\hline
Deep supervision & Enabled \\
\hline
Input channels & 4 \\
\hline
Output channels & 3 \\
\hline
Patch size & $128 \times 160 \times 112$ \\
\hline
Optimizer & AdamW \\
\hline
Learning rate & $1 \times 10^{-4}$ \\
\hline
Weight decay & $1 \times 10^{-5}$ \\
\hline
Batch size & 1 \\
\hline
Gradient accumulation & 2 steps \\
\hline
MedNeXt maximum epochs & 150 \\
\hline
MedNeXt early stopping patience & 30 epochs \\
\hline
Diffusion component & CorrDiff residual error refinement \\
\hline
CorrDiff maximum epochs & 50 \\
\hline
CorrDiff early stopping patience & 30 epochs \\
\hline
CorrDiff diffusion schedule & 200 training timesteps; 25 reverse steps during validation/inference \\
\hline
CorrDiff loss weights & $\lambda_{\mathrm{diff}}=1$, $\lambda_{\mathrm{recon}}=1$, $\lambda_{\mathrm{vq}}=0.25$, $\lambda_{\mathrm{seg}}=1$ \\
\hline
Ensemble size & 5-fold \\
\hline
\end{tabularx}
\endgroup
\caption{\label{tab:training-settings}Core training settings used for CoMNet.}
\end{table*}

\subsubsection{MedNeXt segmentation}

We use MedNeXt as the primary segmentation backbone in CoMNet. The motivation for this comes from ConvNeXt and its adaptation to work well on small clinical datasets. ConvNeXt showed that the convolutional design could be substantially improved when standard ConvNets were updated with larger kernels, depthwise spatial filtering, inverted bottlenecks, residual pathways, and improved normalization \cite{Liu2022}. MedNeXt adapts these ideas to 3D medical image segmentation by placing ConvNeXt-style blocks inside an encoder-decoder architecture \cite{Roy2023}. This is well suited for volumetric brain tumor segmentation from MRI because the model needs to preserve local boundary detail while also using larger anatomical context.

In CoMNet, MedNeXt follows the general encoder-decoder structure used in U-Net-style segmentation models. The encoder progressively reduces the spatial resolution of the input volume while increasing the semantic richness of the learned feature maps. This allows the network to capture larger spatial context across the 3D image. The decoder then restores the feature maps toward the original image resolution and combines them with higher-resolution skip connections from the encoder. This structure is important for glioma segmentation because ET and TC often require fine boundary detail, while WT segmentation depends on recognizing the full extent of the tumor across a larger context window.

The internal MedNeXt blocks make this 3D encoder-decoder more expressive while retaining the advantages of convolutional feature learning. Depthwise 3D convolutions capture spatial patterns within each feature channel, while pointwise convolutions mix information across channels. Each block expands the feature representation, applies nonlinear processing and normalization, and then projects the representation back into a residual pathway. In this way, the block learns a refinement to the current feature representation rather than relearning the full mapping from scratch. This residual design allows MedNeXt to scale to deeper volumetric segmentation models while remaining practical to train using 3D patches. 

We use the updated MedNeXt with GRN to improve feature competition across channels and reduce redundant channel responses \cite{Woo2023}. GRN computes a global summary of each feature channel across the 3D patch and uses this information to modulate channel responses. This allows the model to balance features associated with small tumor subregions, such as enhancing tumor and tumor core, with features representing edema, surrounding anatomy, and broader whole-tumor context. By encouraging more balanced channel responses, GRN may help the network learn a more stable volumetric representation for spatially imbalanced tumor regions.

\subsubsection{Corrective diffusion model}

We adapt CorrDiff as a corrective diffusion post-processing block before ensembling that refines residual errors produced by the MedNeXt segmentation backbone $S_k$. Our design follows the original CorrDiff implementation, which uses diffusion as a corrective model rather than generating the initial segmentation from the image \cite{Li2024}. This distinguishes our approach from many diffusion-based segmentation methods in which the diffusion model serves as the primary mask generator \cite{Ho2020,Kazerouni2023,Wolleb2022,Wu2024}. A diffusion model learns to produce a gradually less noisy $\textbf{z}_{t-1}$ from $\textbf{z}_t$. The model learns this process by randomly sampling a data point $\textbf{z}_0$, a timestep $t$, and noise $\epsilon$, giving a corrupted sample $\textbf{z}_T$. Here, $\textbf{z}_0$ is the error map \textbf{E} from Figure \ref{fig:pipeline} encoded by the VQ-encoder. The forward noising process is a Markov chain that corrupts data $\textbf{z}_0\sim q(\textbf{z}_0)$ over $T$ timesteps by adding Gaussian noise $\epsilon$ at time $t$, giving $\textbf{z}_0\to\textbf{z}_T$:
\begin{equation}
q(\textbf{z}_{1:T}|\textbf{z}_0)=\prod_{t=1}^Tq(\textbf{z}_t|\textbf{z}_{t-1})
\end{equation}
\begin{equation}
q(\textbf{z}_t|\textbf{z}_{t-1})=\mathcal{N}(\textbf{z}_t;\sqrt{1-\beta_t} \textbf{z}_{t-1},\beta_t \textbf{I})),
\end{equation}
where $\mathcal{N}$ is the Gaussian distribution, \textbf{I} is the identity matrix, $\beta_t$ is a constant between 0 and 1, and $\sqrt{\beta_t}$ is $\sigma_t$. The reverse diffusion process $\textbf{z}_T\to\textbf{z}_0$ is a Markov chain process expressed by:
\begin{equation}
p_\theta(\textbf{z}_{t-1}|\textbf{z}_t, \tau_\theta(\textbf{X}))=\mathcal{N}(\textbf{z}_{t-1};\textbf{$\mu$}_\theta(\textbf{z}_{t},t,\tau_\theta(\textbf{X})),\Bar{\beta_t}\textbf{I})
\end{equation}
\begin{equation}
p_\theta(\textbf{z}_{0:T-1}|\textbf{z}_T, \tau_\theta(\textbf{X}))=\prod_{t=1}^T p_\theta(\textbf{z}_{t-1}|\textbf{z}_t,\tau_\theta(\textbf{X})),
\end{equation}
where \textbf{X} is the MR image, $\textbf{$\mu$}_\theta(\textbf{z}_{t},t,\tau_\theta(\textbf{X}))$ is the mean and $\Bar{\beta_t}\textbf{I}$ is the variance of the reverse diffusion process. After the reverse process, $\textbf{z}_0$ is then decoded to obtain the reconstructed error map $\hat{E}$. The final corrected mask $\hat{M}$ is generated by integrating $\hat{E}$ into $S$; this process is applied fold-wise before the final ensemble. 

After initial training of MedNeXt, we train the corrective diffusion model on each dataset while keeping the MedNeXt model weights frozen. The corrective diffusion model receives the MRI volume, the MedNeXt coarse probabilities, the MedNeXt coarse logits, and an entropy-based uncertainty map. The correction target is the residual between the binary tumor-region target and the MedNeXt probability map. Rather than diffusing this residual directly in the full image space, we first encode the residual with a 3D vector quantized variational autoencoder (VQ-VAE). The latent diffusion model then learns to denoise this compact residual representation while conditioned on the MRI and MedNeXt output. At inference, CorrDiff samples a latent residual correction, decodes it into a correction to the MedNeXt logits, gates that correction using the coarse probabilities and uncertainty map, and adds the gated correction back to the MedNeXt logits before final sigmoid thresholding. CorrDiff is trained with a composite objective function that combines diffusion denoising, residual reconstruction, vector-quantization regularization, and final segmentation supervision:
\begin{equation}
L_{\mathrm{CorrDiff}} =
\lambda_{\mathrm{diff}}L_{\mathrm{diff}} +
\lambda_{\mathrm{recon}}L_{\mathrm{recon}} +
\lambda_{\mathrm{vq}}L_{\mathrm{vq}} +
\lambda_{\mathrm{seg}}L_{\mathrm{seg}}
\label{eq:corrdiff-loss}
\end{equation}
The diffusion term $L_{\mathrm{diff}}$ is an MSE noise-prediction loss in the latent space. $L_{\mathrm{recon}}$ is an L1 loss for reconstructing the target residual $r = y - \sigma(\ell_{\mathrm{coarse}})$. $L_{\mathrm{vq}}$ is the VQ-VAE codebook and commitment loss, and $L_{\mathrm{seg}}$ combines binary cross-entropy with Dice loss on the corrected logits $\ell_{\mathrm{final}} = \ell_{\mathrm{coarse}} + \Delta \ell$. We use $\lambda_{\mathrm{diff}} = 1$, $\lambda_{\mathrm{recon}} = 1$, $\lambda_{\mathrm{vq}} = 0.25$, and $\lambda_{\mathrm{seg}} = 1$, encouraging CorrDiff to learn corrections that are plausible in the latent diffusion space and beneficial for the final tumor-region segmentation.

\subsubsection{Fold-wise ensemble}

We use five-fold cross-validation for training, each fold trains a unique MedNeXt model fit on a different training split and validated on a unique held-out split. These fold-specific models can learn slightly different decision boundaries. In our proposed framework, each fold-specific model first produces region-wise logits for ET, TC, and WT. The outputs are treated as three independent binary prediction channels. For fold $k \in \{1,\ldots,K\}$, region channel $c \in \{\mathrm{ET},\mathrm{TC},\mathrm{WT}\}$, and voxel $v$, let

\begin{equation}
z_{k,c}(v) = f_{\theta_k}(\textbf{X})_c(v)
\end{equation}
denote the MedNeXt logit $z$ for fold $k$. The corresponding coarse probability is then:
\begin{equation}
p_{k,c}(v) = \sigma(z_{k,c}(v)),
\end{equation}
where $\sigma(\cdot)$ is the sigmoid function. Each fold prediction is then passed through the corrective diffusion model and predicts a probability-space residual $r$ conditioned on the MRI volume \textbf{X}, the coarse logit maps $\mathbf{z}_k$, the coarse probability maps $\mathbf{p}_k = \sigma(\mathbf{z}_k)$, and an uncertainty map $\mathbf{u}_k$. The ensemble probability map is then computed by averaging the corrected probabilities voxel-wise across the folds:
\begin{equation}
\bar{p}_{c}(v)
=
\frac{1}{K}
\sum_{k=1}^{K}
\hat{p}_{k,c}(v).
\end{equation}
The final binary segmentation for each region is obtained by applying a fixed threshold $\tau=0.5$ independently to each channel:
\begin{equation}
\bar{M}_c(v)
=
\mathbb{I}\left[\bar{p}_{c}(v) > \tau\right],
\qquad \tau = 0.5.
\end{equation}
Meaning, a voxel is assigned to region $r$ only when the mean corrected probability across folds exceeds 0.5. Averaging is performed before thresholding because it preserves the continuous confidence information from each fold. This differs from majority voting on binary masks, which would discard whether a fold predicted a voxel with high confidence or only marginally above threshold. Voxels consistently assigned high probability across folds remain high confidence after averaging, while unstable fold-specific predictions are softened toward intermediate probabilities and may fall below the final threshold. This produces the final segmentation prediction, shown as $\bar{M}$ in Figure~\ref{fig:pipeline}, using information from all independently trained fold models while reducing variance from any single fold.

\subsection{Loss functions}

We train each segmentation model with a region overlap aware loss rather than a pure voxel-wise loss. This is useful because tumor regions occupy a small fraction of the full 3D brain volume. Dice-style objective functions directly optimize region overlap. For region channel $c$, the soft Dice score is
\begin{equation}
\mathrm{DSC}_{c} =
\frac{2\sum_i p_{c,i}y_{c,i} + \epsilon}
{\sum_i p_{c,i} + \sum_i y_{c,i} + \epsilon},
\end{equation}
where $p_{c,i}$ is the predicted probability for voxel $i$, $y_{c,i}$ is the binary ground-truth label, and $\epsilon$ is a small constant for numerical stability. The standard Dice loss is defined as the average Dice loss across ET, TC, and WT:
\begin{equation}
L_{\mathrm{Dice}} =
1 - \frac{1}{3}\sum_{c \in \{ET,TC,WT\}}\mathrm{DSC}_{c}.
\end{equation}
We train CoMNet using a DSC++ variant, which replaces the standard Dice denominator with a squared probability term:
\begin{equation}
\mathrm{DSC}^{++}_{c} =
\frac{2\sum_i p_{c,i}y_{c,i} + \epsilon}
{\sum_i p_{c,i}^{2} + \sum_i y_{c,i}^{2} + \epsilon}.
\end{equation}
Our final loss combines DSC++ with binary cross-entropy:
\begin{equation}
L_{\mathrm{MedNeXt}} =
L_{\mathrm{DSC}^{++}} +
L_{\mathrm{BCE}}.
\end{equation}
This allows to keep the emphasis on overlap while adding voxel-wise supervision, however, we still evaluate using the Dice similarity coefficient as the standard overlap metric for ET, TC, and WT segmentation. We hold the training configuration consistent across SegResNet and MedNeXt experiments to ensure the performance differences can be interpreted primarily as differences between model designs and the CoMNet framework.

\subsection{Evaluation metrics}

Segmentation performance is evaluated separately for ET, TC, and WT. For each region, the model prediction is converted into a binary mask and compared with the corresponding binary ground-truth mask. Let $P_c$ denote the predicted voxel set for region channel $c$, and let $G_c$ denote the ground-truth voxel set. From these masks, we compute the number of true positives ($TP$), false positives ($FP$), true negatives ($TN$), and false negatives ($FN$). The Dice similarity coefficient is the primary metric:
\begin{equation}
\mathrm{DSC}_c = \frac{2|P_c \cap G_r|}{|P_c| + |G_c|} = \frac{2TP}{2TP + FP + FN}.
\end{equation}
Dice measures the overlap between the predicted and ground-truth tumor region, with 1 indicating perfect overlap and 0 indicating no overlap. We also compute the region-wise average:
\begin{equation}
\mathrm{DSC}_{\mathrm{avg}} =
\frac{1}{3}\left(\mathrm{DSC}_{ET} + \mathrm{DSC}_{TC} + \mathrm{DSC}_{WT}\right).
\end{equation}
This average is used for fold-level model selection and for the overall comparison between SegResNet, MedNeXt, and CoMNet.

\subsection{Implementation details}

All models are trained on image patches cropped to a fixed size of $128 \times 160 \times 112$. Hyperparameter values are selected empirically, and changes to these settings can substantially affect model performance. The pipeline is implemented in Python, with PyTorch used for model training. TorchIO is used for medical image preprocessing and augmentation, MONAI utilities support medical imaging workflows and baseline loss functions, and NumPy and SciPy are used for numerical operations and metric computation. CorrDiff is implemented as a VQ-VAE- and DDPM-based residual refinement module. All network training and experiments are performed on an NVIDIA V100 GPU with 16 GB of VRAM. All models are trained with AdamW using a learning rate of $1 \times 10^{-4}$ and a weight decay of $1 \times 10^{-5}$. We use a batch size of 1 with gradient accumulation over 2 steps, resulting in an effective batch size of 2. Training includes a 5-epoch warmup followed by cosine annealing, a maximum of 150 epochs, early stopping with a patience of 30 epochs, mixed-precision training, and gradient clipping with a maximum norm of 1.0. CorrDiff is trained as a second-stage post-processing module after MedNeXt training is complete, with the MedNeXt weights frozen. Core training settings are summarized in Table \ref{tab:training-settings}.

\section{Experiment results}

\subsection{Quantitative analysis}

\begin{figure*}[!h]
\centering
\includegraphics[width=0.9012\textwidth]{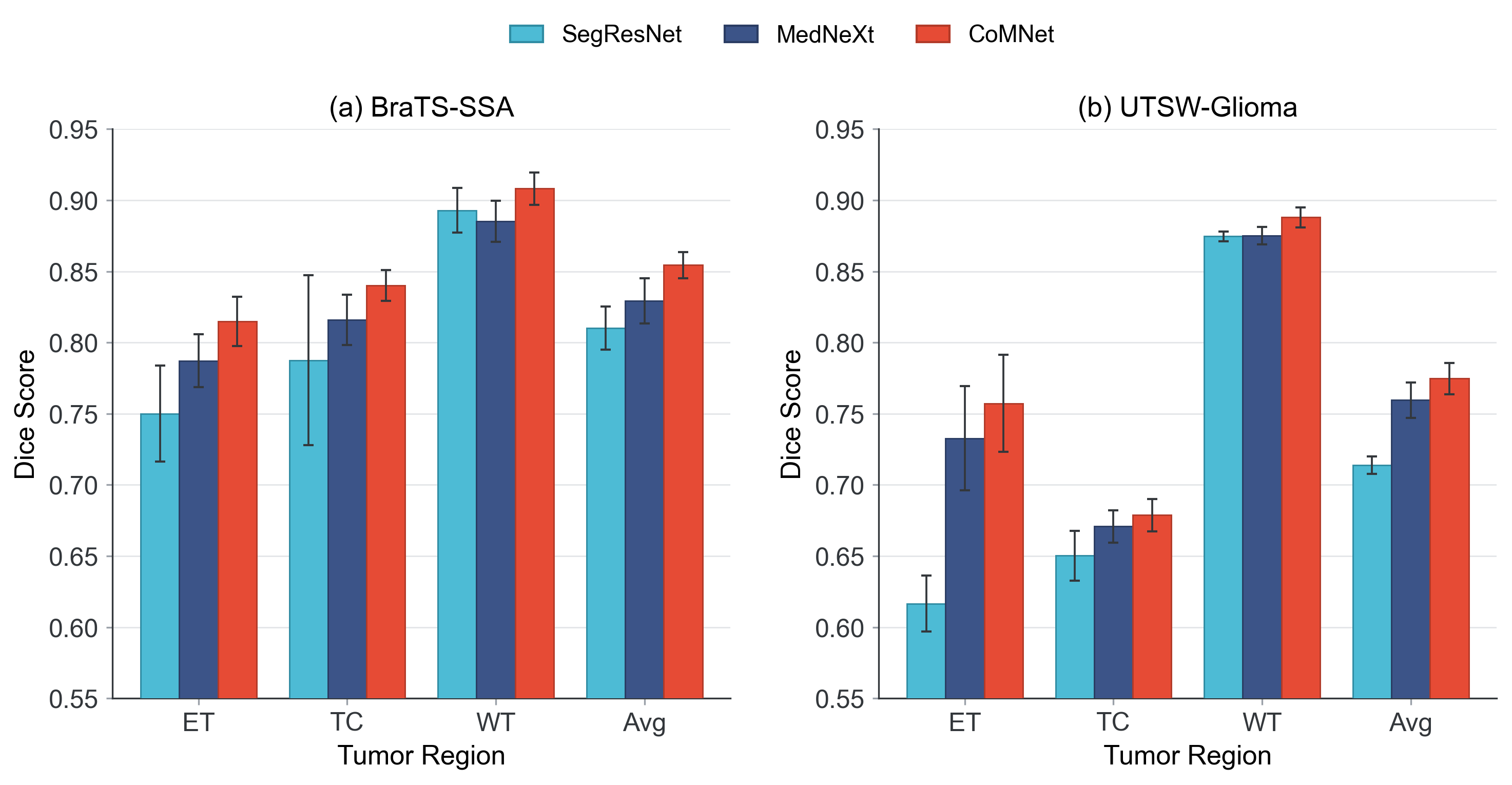}
\caption{\label{fig:heldout-test-performance-by-tumor-region}Held-Out Test Performance by Tumor Region. DSC on the held-out test sets of \textbf{(a)} BraTS-SSA $(n = 12)$ and \textbf{(b)} UTSW-Glioma $(n = 125)$, for the ET, TC, WT, and their average (Avg). SegResNet and MedNeXt are reported as the mean $\pm$ standard deviation across the five cross-validation folds; CoMNet (proposed) is reported as the five-fold ensemble with CorrDiff refinement. The standard deviation for CoMNet is reported as the five per-fold, CorrDiff-refined predictions before the ensemble, drawn at the ensemble final value to indicate fold-level stability. The training loss is selected per cohort (DSC++CE for BraTS-SSA, Dice for UTSW-Glioma; see Figure \ref{fig:loss_ablation}).}
\end{figure*}

As shown in Table \ref{tab:all-results} and Figure \ref{fig:heldout-test-performance-by-tumor-region}, the proposed framework outperforms SegResNet and standalone MedNeXt on the held-out test sets of both BraTS-SSA and UTSW-Glioma. Across both cohorts, CoMNet achieves the highest Dice score for enhancing tumor (ET), tumor core (TC), whole tumor (WT), and the region-wise average. These results indicate that combining a modern convolutional segmentation backbone with CorrDiff residual refinement and fold-level probability ensembling improves segmentation accuracy across heterogeneous glioma MRI datasets.

On the BraTS-SSA held-out test set, CoMNet trained with DSC++CE achieves DSC scores of 0.8150, 0.8400, and 0.9082 for ET, TC, and WT, respectively, with an average DSC of 0.8544. Compared with SegResNet, this corresponds to absolute DSC improvements of 0.0649 for ET, 0.0524 for TC, 0.0153 for WT, and 0.0442 on average. Compared with standalone MedNeXt, CoMNet improves ET, TC, WT, and average DSC by 0.0278, 0.0242, 0.0230, and 0.0250, respectively. These results show that CoMNet improves not only the smaller ET and TC regions, but also the larger WT region, where baseline performance is already relatively high.

On the UTSW-Glioma held-out test set, CoMNet trained with Dice loss achieves DSC scores of 0.7579, 0.6787, and 0.8879 for ET, TC, and WT, respectively, with an average DSC of 0.7749. Compared with SegResNet, CoMNet improves ET, TC, WT, and average DSC by 0.1413, 0.0285, 0.0132, and 0.0611, respectively. The largest improvement is observed for ET, where CoMNet increases DSC by 14.1 percentage points over SegResNet. Compared with standalone MedNeXt, CoMNet improves ET, TC, WT, and average DSC by 0.0252, 0.0078, 0.0128, and 0.0154, respectively.

\begin{figure*}[!t]
\centering
\includegraphics[width=0.86\textwidth]{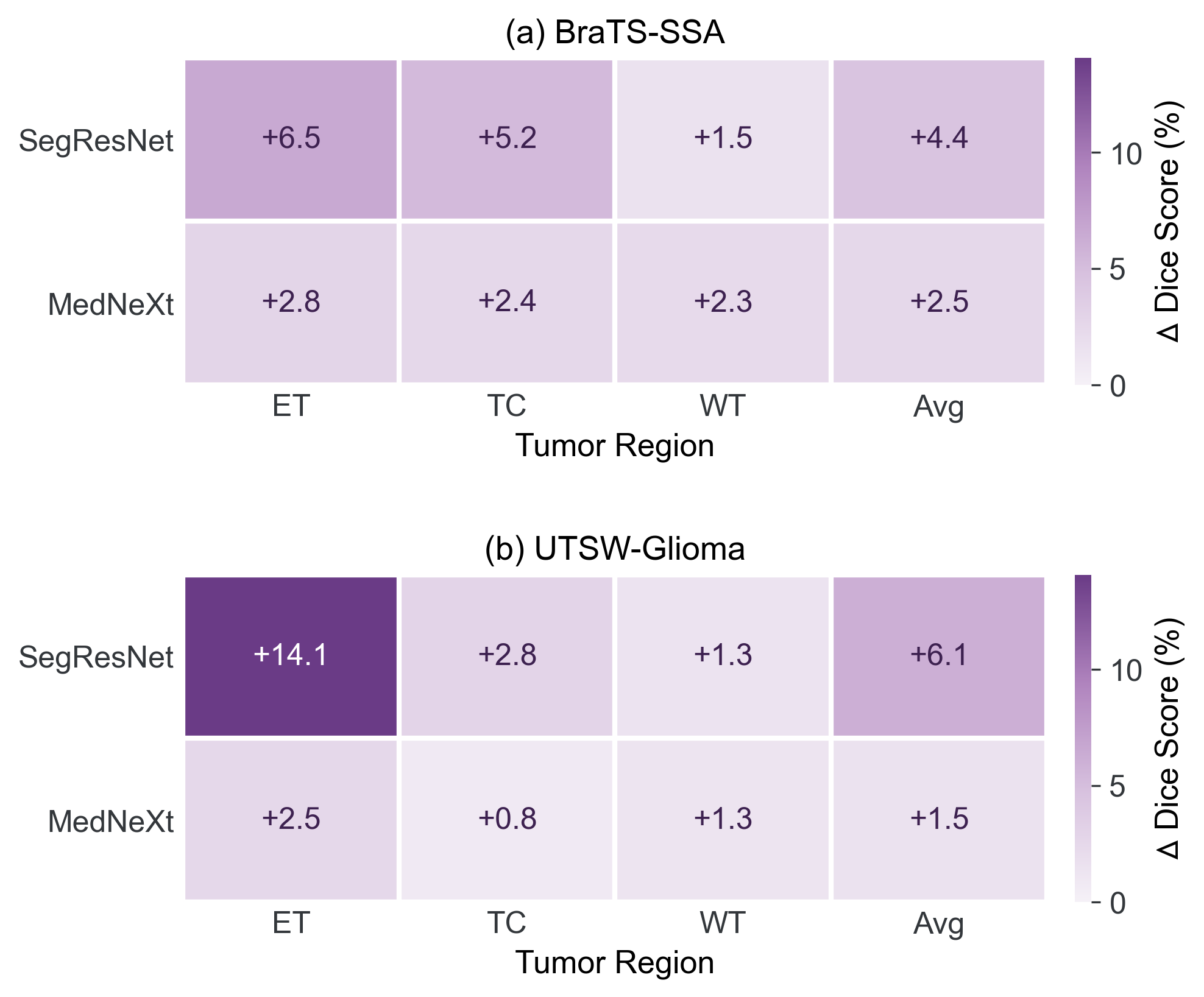}
\caption{\label{fig:improvement_heatmap}CoMNet Performance Improvement by Dataset and Tumor Region. Absolute change in Dice similarity coefficient ($\Delta$ DSC) of CoMNet relative to SegResNet and MedNeXt on the held-out test sets, shown separately for \textbf{(a)} BraTS-SSA and \textbf{(b)} UTSW-Glioma. Each cell is CoMNet minus the baseline's five-fold mean DSC for the enhancing tumor (ET), tumor core (TC), whole tumor (WT), and their average (Avg). Positive values represent an improvement on each tumor region against the comparing baseline. The largest single gain (+14.1 points) is for the enhancing tumor over SegResNet on UTSW-Glioma.}
\end{figure*}

Figure \ref{fig:improvement_heatmap} further summarizes these performance gains by showing the absolute DSC improvement of CoMNet over both baselines for each dataset and tumor region. All values are positive, demonstrating that CoMNet improves performance across every region and cohort. The improvement is especially notable for ET on UTSW-Glioma against SegResNet, which suggests that the proposed refinement and ensembling strategy is useful for spatially limited tumor regions that are more sensitive to small false-positive and false-negative errors.

\begin{figure*}[!t]
\centering
\includegraphics[width=0.9012\textwidth]{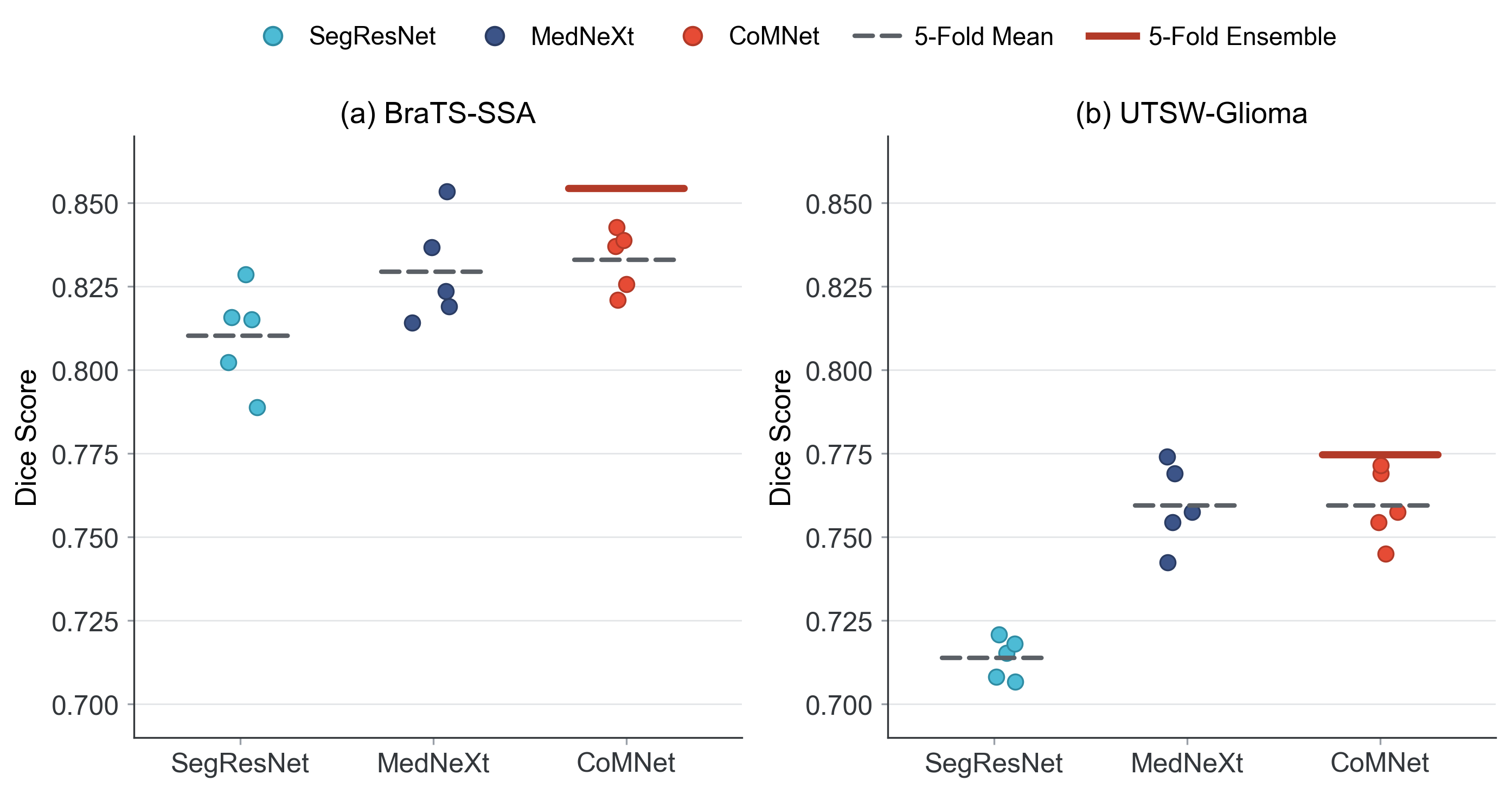}
\caption{\label{fig:perfold_test_dots}Distribution of Held-Out Test Dice Scores Across Cross-Validation Folds. Per-fold held-out test Dice similarity coefficient (averaged over ET, TC, and WT) for \textbf{(a)} BraTS-SSA and \textbf{(b)} UTSW-Glioma. For each model, dots represent the five individual cross-validation folds and the dashed line their five-fold mean. For the baseline models, each dot is a single trained model. For CoMNet, each dot is the corresponding fold's MedNeXt model with CorrDiff refinement prior to ensembling; the solid red line is post ensemble. The proposed framework exceeds the mean of CoMNet's individual folds, isolating the performance increase attributable to ensembling.}
\end{figure*}

Figure \ref{fig:perfold_test_dots} shows the distribution of held-out average DSC across the five cross-validation folds. For SegResNet and MedNeXt, each point represents an independently trained fold model. For CoMNet, the points represent the CorrDiff-refined fold predictions before ensembling, while the solid red line represents the deployed five-fold probability ensemble. On both datasets, the CoMNet ensemble exceeds the mean of its own fold-level predictions and also exceeds the baseline fold results. This supports the use of probability-level ensembling, since averaging corrected probability maps before thresholding reduces fold-specific variance and produces a more reliable final segmentation on multi-site datasets.

\begin{figure*}[!t]
\centering
\includegraphics[width=\textwidth]{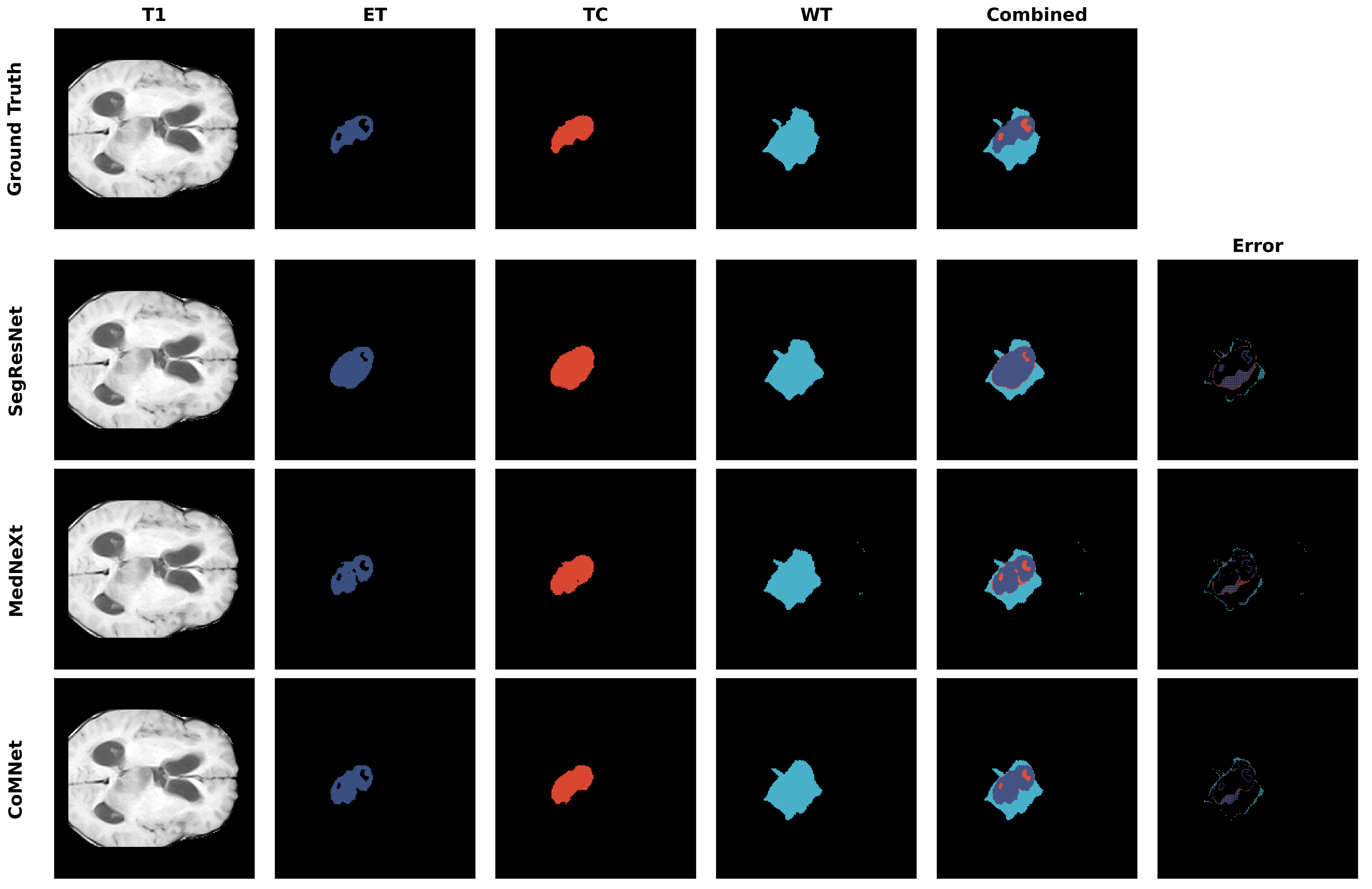}
\caption{\label{fig:prediction}Visualization of Segmentation Results on a Held-Out BraTS-SSA Test Patient. The error map represents every voxel where a prediction disagrees with the ground truth; a more accurate model produces a sparser error cloud.
}
\end{figure*}

\subsection{Qualitative analysis}

Figure \ref{fig:prediction} shows a qualitative comparison between the ground truth, SegResNet, baseline MedNeXt, and CoMNet on a held-out BraTS-SSA test patient. The visualization includes the T1 image, region-wise ET, TC, and WT masks, the combined segmentation, and an error map showing disagreement between the predicted mask and the ground truth. Compared with SegResNet, CoMNet produces a segmentation that more closely follows the ground-truth tumor structure, particularly in the ET and combined tumor regions. Compared with MedNeXt, CoMNet reduces isolated prediction errors and provides a cleaner WT segmentation. The error map is also sparser for CoMNet, indicating fewer voxel-level disagreements with the manual annotation. This visual pattern is consistent with the quantitative results, where CoMNet achieves the highest DSC across all tumor regions on the BraTS-SSA held-out test set.

\subsection{Ablation study}

We evaluate the contribution of the major components of CoMNet using the MedNeXt baseline, the MedNeXt + CorrDiff refinement stage, the full CoMNet ensemble, and the training loss ablation. As shown in Table \ref{tab:all-results}, adding CorrDiff to MedNeXt provides a modest improvement on BraTS-SSA, increasing the average DSC from 0.8294 to 0.8331 before the full ensemble is applied. The complete CoMNet pipeline further increases the average DSC to 0.8544. On UTSW-Glioma, MedNeXt + CorrDiff alone obtains a similar average DSC to standalone MedNeXt, while the full CoMNet ensemble increases the average DSC from 0.7595 to 0.7749. These results indicate that CorrDiff refinement and fold-level ensembling are complementary, with the larger gain coming from the deployed ensemble prediction, while corrective diffusion reduces variance between folds.

\begin{figure*}[!h]
\centering
\includegraphics[width=0.9745\textwidth]{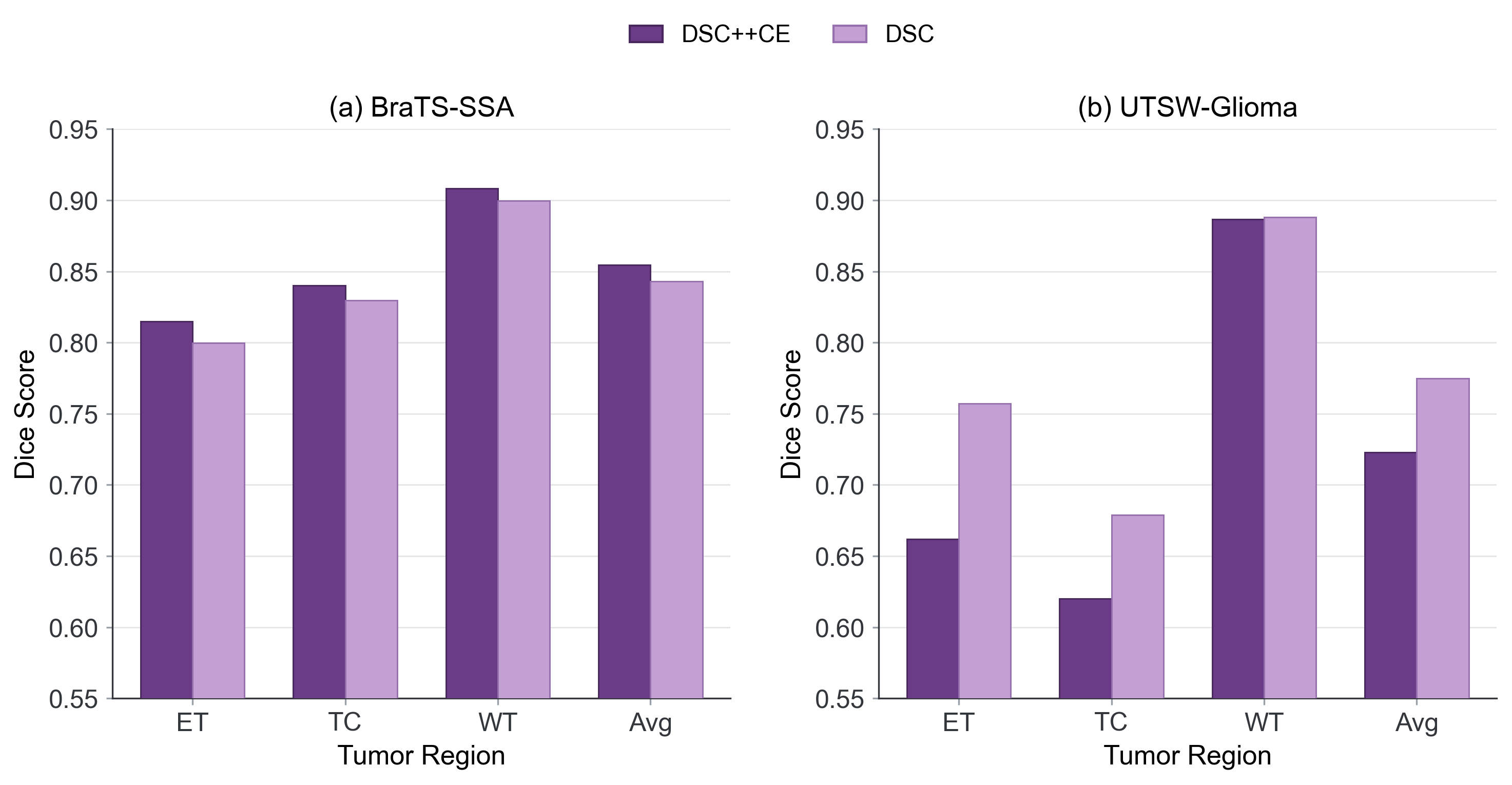}
\caption{\label{fig:loss_ablation}Effect of Training Loss on Held-Out Test Set Performance by Dataset. Held-out test Dice similarity coefficient (DSC) CoMNet trained with the Dice++ and cross-entropy loss (DSC++CE) versus DSC by tumor region for \textbf{(a)} BraTS-SSA and \textbf{(b)} UTSW-Glioma. DSC++CE loss generalizes best on the smaller BraTS-SSA cohort, whereas DSC loss is best on the larger UTSW-Glioma cohort. The better loss per cohort is used throughout the remaining analyses.}
\end{figure*}

Figure \ref{fig:loss_ablation} shows the effect of training loss on CoMNet performance. On BraTS-SSA, DSC++CE achieves the highest performance, improving the average DSC from 0.8430 with Dice loss to 0.8544. In contrast, standard Dice loss performs better on UTSW-Glioma, increasing the average DSC from 0.7229 with DSC++CE to 0.7749. Therefore, DSC++CE is used for BraTS-SSA and Dice loss is used for UTSW-Glioma in the final reported CoMNet configuration.

\begin{table*}[!t]
\centering
\begingroup
\renewcommand{\arraystretch}{1.25}
\setlength{\tabcolsep}{8pt}
\renewcommand{\tabularxcolumn}[1]{m{#1}}
\begin{tabularx}{\textwidth}{|>{\centering\arraybackslash}X|
>{\centering\arraybackslash}X|
>{\centering\arraybackslash}X|
>{\centering\arraybackslash}X|
>{\centering\arraybackslash}X|
>{\centering\arraybackslash}X|
>{\centering\arraybackslash}X|}
\hline
\rowcolor{gray!15}
\textbf{Dataset} & \textbf{Model} & \textbf{Loss} & \textbf{ET} & \textbf{TC} & \textbf{WT} & \textbf{Average} \\
\hline
BraTS-SSA & SegResNet & DSC & $0.7501$ & $0.7876$ & $0.8929$ & $0.8102$ \\
\hline
BraTS-SSA & MedNeXt & DSC & $0.7872$ & $0.8158$ & $0.8852$ & $0.8294$ \\
\hline
BraTS-SSA & MedNeXt + CorrDiff & DSC++CE & $0.7940$ & $0.8132$ & $0.8920$ & $0.8331$ \\
\hline
BraTS-SSA & CoMNet (proposed) & DSC & $0.7996$ & $0.8297$ & $0.8956$ & $0.8430$ \\
\hline
\textbf{BraTS-SSA} & \textbf{CoMNet (proposed)} & \textbf{DSC++CE} & $\mathbf{0.8150}$ & $\mathbf{0.8400}$ & $\mathbf{0.9082}$ & $\mathbf{0.8544}$ \\
\hline
UTSW-Glioma & SegResNet & DSC & $0.6166$ & $0.6502$ & $0.8747$ & $0.7138$ \\
\hline
UTSW-Glioma & MedNeXt & DSC & $0.7327$ & $0.6709$ & $0.8751$ & $0.7595$ \\
\hline
UTSW-Glioma & MedNeXt + CorrDiff & DSC & $0.7342$ & $0.6709$ & $0.8735$ & $0.7595$ \\
\hline
UTSW-Glioma & CoMNet (proposed) & DSC++CE & $0.6620$ & $0.6200$ & $0.8867$ & $0.7229$ \\
\hline
\textbf{UTSW-Glioma} & \textbf{CoMNet (proposed)} & \textbf{DSC} & $\mathbf{0.7579}$ & $\mathbf{0.6787}$ & $\mathbf{0.8879}$ & $\mathbf{0.7749}$ \\
\hline
\end{tabularx}
\endgroup
\caption{\label{tab:all-results}Held-out test-set segmentation performance on BraTS-SSA ($n=12$) and UTSW-Glioma ($n=125$). Entries are Dice similarity coefficient (DSC) scores. The SegResNet and MedNeXt baselines and the MedNeXt\,+\,CorrDiff ablation are reported as the mean over the five cross-validation folds. CoMNet is reported as a five-fold probability ensemble with CorrDiff refinement. The highest DSC on every tumor region within each dataset are shown in bold.}
\end{table*}

\section{Discussion}

In this study, we propose CoMNet to improve brain tumor segmentation by refining residual errors from a strong MedNeXt backbone and combining fold-specific predictions through probability ensembling. Rather than replacing the segmentation network, CorrDiff is used as a correction stage that learns from the mismatch between the initial probability maps and the manual annotation. The experimental results show that this strategy improves held-out test set DSC on both BraTS-SSA and UTSW-Glioma, with CoMNet achieving the highest ET, TC, WT, and average Dice scores across both cohorts.

The improvement is especially meaningful for ET, which is often the most difficult region because it occupies a smaller volume and is sensitive to small false-positive or false-negative errors. On UTSW-Glioma, CoMNet improves ET DSC by 14.1\% over SegResNet and 2.5\% over MedNeXt. On BraTS-SSA, CoMNet improves the average DSC by 4.4\% over SegResNet and 2.5\% over MedNeXt. These results suggest that the combination of corrective diffusion and fold-level ensembling can improve segmentation accuracy across both a smaller underrepresented cohort and a larger multi-site clinical dataset.

In the held-out BraTS-SSA example, CoMNet produces a segmentation that more closely follows the ground-truth tumor structure and yields a sparser error map than the baseline models. This indicates that the proposed framework can reduce local segmentation disagreement, particularly around tumor boundaries and smaller subregions. Since CorrDiff operates on residual prediction errors, it performs a focused refinement task rather than generating the entire segmentation mask from the image.

Our proposed method also has practical value as a modular framework. Because the correction stage is applied after the backbone prediction, it can potentially be adapted to other segmentation networks that output probabilistic masks. In addition, probability-level ensembling reduces the influence of any single fold-specific model and produces a more stable consensus prediction. This is especially important for multi-site MRI datasets, where scanner differences, acquisition variability, and patient-level heterogeneity can lead to unstable model predictions.

There are also several limitations. First, the choice of loss function is dataset dependent. Although DSC++CE performs best on the smaller BraTS-SSA cohort, it performs worse on UTSW-Glioma, where the DSC++CE CoMNet configuration achieves an average DSC of 0.7229, below the Dice-trained MedNeXt baseline average of 0.7595. This shows that the compound loss does not universally improve performance and may interact differently with dataset size, tumor-region distribution, and cohort heterogeneity. For this reason, loss-function selection remains an important part of model development.

Second, CorrDiff alone does not improve the mean average Dice on UTSW-Glioma compared with standalone MedNeXt. Both MedNeXt and MedNeXt + CorrDiff achieve an average DSC of approximately 0.7595 before the full ensemble is applied. However, CorrDiff slightly reduces the fold-level standard deviation of average DSC, from 0.0125 to 0.0109, suggesting that its benefit on UTSW-Glioma is more related to prediction stability than mean performance improvement. This robustness is still valuable in multi-site segmentation, but it indicates that CorrDiff refinement may not always produce a direct Dice increase without ensembling.

Finally, the proposed framework increases computational cost because it requires training multiple fold-specific models and applying a second-stage corrective diffusion block. Future work should evaluate faster sampling strategies, additional loss formulations, and broader external validation across more datasets. Further analysis with boundary-based metrics such as HD95 would also help determine whether the visual boundary improvements translate into stronger surface-distance performance.

\section{Conclusion}

In this work, we proposed CoMNet, a MedNeXt-CorrDiff framework for multi-site MRI brain tumor segmentation. The framework first uses MedNeXt as the primary 3D segmentation backbone to generate initial tumor probability maps. CorrDiff is then applied as a post-processing prediction refinement method to model residual segmentation errors, and the corrected fold-specific predictions are combined through probability-level ensembling to produce the final ET, TC, and WT masks. We evaluated CoMNet on the BraTS-SSA and UTSW-Glioma datasets using held-out test sets unseen during training. Experimental results show that CoMNet achieves the highest Dice scores compared with SegResNet and standalone MedNeXt across all tumor regions and both datasets. These results demonstrate that corrective diffusion and fold-level ensembling can improve segmentation accuracy and stability across heterogeneous MRI cohorts. Future work will focus on improving loss-function generalization, reducing diffusion inference cost, and validating the framework on additional external brain tumor datasets.

\section*{Acknowledgements}

This study is supported in part by the National Institutes of Health (NIH) under grant R01 EB020683.

\section*{Additional information}

\subsection*{Competing interests}

The authors declare no competing interests.

\subsection*{Data availability}

The UTSW-Glioma dataset analyzed in this study is publicly available through The Cancer Imaging Archive (TCIA) under the UTSW-Glioma collection, subject to the applicable TCIA access requirements. The BraTS-SSA dataset analyzed in this study is publicly available through TCIA under DOI: 10.7937/v8h6-8x67.

\subsection*{Code availability}

The code supporting this study is not publicly available at this time, as it is part of an ongoing research project. Additional implementation details are available from the corresponding author upon reasonable request.

\bibliography{references}

\end{document}